\documentclass[preprint,12pt]{elsarticle}
\usepackage[dvipsnames]{xcolor}
\usepackage[numbers]{natbib}
\newcommand{\green}[1]{\textcolor{Green}{#1}}
\newcommand{\red}[1]{\textcolor{Red}{#1}}

\usepackage{times}
\usepackage{epsfig}
\usepackage{graphicx,subcaption}
\usepackage{amsmath}
\usepackage{amssymb}
\usepackage{lipsum}
\usepackage{mathtools, nccmath}

\DeclarePairedDelimiter{\nint}\lfloor\rceil

\usepackage[ruled,vlined]{algorithm2e}

\newcommand{\lce}{\mathcal{L}_{\text{CE}}}
\newcommand{\lmc}{\mathcal{L}_{\text{MC}}}

\begin{document}


\title{Learning Discriminative Representations for Multi-Label Image Recognition}

\author [lb1]{Mohammed Hassanin}
\address[lb1]{UNSW Canberra, Australia. E-mail: m.hassanin@unsw.edu.au}

\author [lb2]{Ibrahim Radwan}
\address[lb2]{University of Canberra. E-mail: ibrahim.radwan@canberra.edu.au}

\author [lb15]{Salman Khan}
\address[lb15]{IIAT, UAE. E-mail: salman.khan@inceptionai.org}

\author [lb3]{Murat Tahtali}
\address[lb3]{UNSW Canberra, Australia. E-mail: murat.tahtali@adfa.edu.au}

\cortext[cor1]{Corresponding author: Mohammed Hassanin}

\begin{abstract}
 Multi-label recognition is a fundamental, and yet is a challenging task in computer vision. Recently, deep learning models have achieved great progress towards learning discriminative features from input images. However, conventional approaches are unable to model the inter-class discrepancies among features in multi-label images, since they are designed to work for image-level feature discrimination. In this paper, we propose a unified deep network to learn discriminative features for the multi-label task. Given a multi-label image, the proposed method first disentangles features corresponding to different classes. Then, it discriminates between these classes via increasing the inter-class distance while decreasing the intra-class differences in the output space. By regularizing the whole network with the proposed loss, the performance of applying the well-known ResNet-101 is improved significantly. Extensive experiments have been performed on COCO-2014, VOC2007 and VOC2012 datasets, which demonstrate that the proposed method outperforms state-of-the-art approaches by a significant margin of $3.5\%$ on large-scale COCO dataset. Moreover, analysis of the discriminative feature learning approach shows that it can be plugged into various types of multi-label methods as a general module.
\end{abstract}
\maketitle

\section{Introduction}
\begin{figure}[t]{
    \centering
       \includegraphics[width=\linewidth]{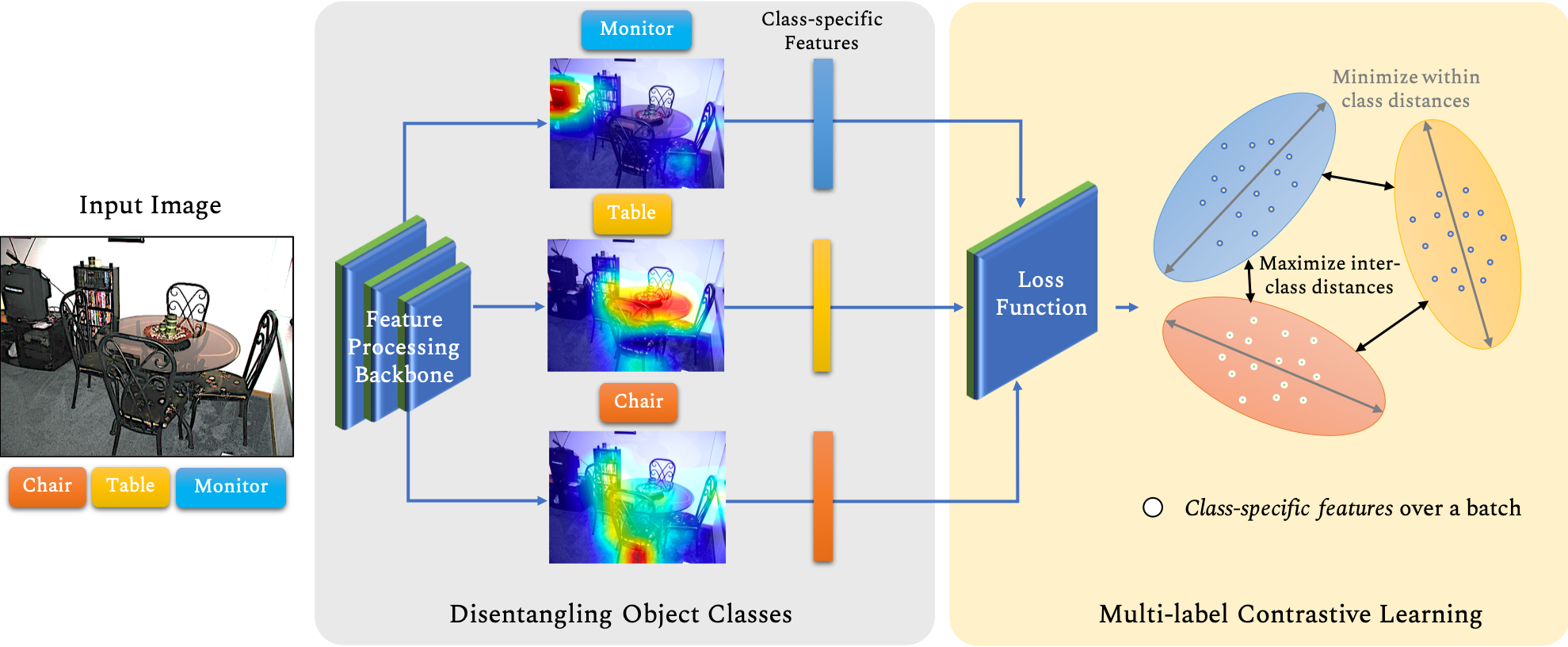}}
   \caption{The input image is fed into the system to disentangle the class-specific features. Then, the proposed loss function is applied to discriminate the classes' features.}
  \label{fig:intuition} 
\end{figure}
 
 The objective of Multi-label recognition task is to predict the presence of multiple objects in the input image. This makes it a fundamental basis in building robust visual recognition systems, because it is more realistic than the traditional classification, especially in application domains such as medical diagnosis recognition \cite{medical_class}, person re-identification \cite{reid_1,reid_2,reid_3} and image tagging \cite{tagging_1,tagging_2, ijcai2018, shen2017multilabel}. Since the beginning of Convolutional Neural Networks (CNN), various paradigms of visual recognition have achieved significant improvement in image classification \cite{Resnet_2016,class_2,class_3}, attribute recognition \cite{att_1,att_2,att_3,att_4,att_5}, object detection \cite{YOLO_redmon2016,FasterR-CNN_2015,SSDliu2016} and image segmentation \cite{Mask-R-CNN_2017_ICCV,giou,my_aff}. However, multi-label recognition has received less attention from the research community because it is challenging to learn discriminative representations for different objects present in a single image. 
 
 %
%

In traditional visual recognition, learning discriminative representations increased the ability of the CNN based methods to improve the performance and achieve better accuracy \cite{centerloss,discriminative,l2_discriminative}. Learning these representations is pursued by learning a mapping function between the input and output in a one-to-one relationship, where the corresponding class is dominating the learning representations. However, in multi-label recognition, this function is required to learn many-to-many mapping between input and output spaces. This creates more burdens on the model to discriminate between the features of the different classes, which may be present in the same input. In this paper, we firstly tackle this problem by proposing an object parser to separate the image features into class-specific representations. This parser disentangles the interference of the classes in the feature space, which simplifies the problem of learning discriminative~representations.
 
We note that disentangling the objects in the feature space will not separate them explicitly as they still can interfere in the output classification space. This is due to the combinatorial nature of the output space in the multi-label recognition problem. In order to tackle this problem, there is a need to minimise the intra-class variations and maximise the inter-class separation of the objects in the output space while training the network. Several objective functions have been proposed for this purpose, such as center loss \cite{centerloss}, triplet loss \cite{triplet1,triplet2}, contrastive loss \cite{contrastive_1,contrastive_2}. Inspired by their success in face recognition task, and based on the above observation, we introduce a multi-label contrastive loss as an objective function to learn discriminative representations of the objects from different classes.
%
 %

The proposed framework in this paper combines two main steps (as illustrated in Figure \ref{fig:intuition}): Firstly, during the training process, we disentangle the classes' representations in the feature manifold in order to simplify the many-to-many mapping into one-to-one relationship between the features and their classes. In this step, the bounding-boxes' annotations are used to separate the spatial features according to their classes. Secondly, a multi-label contrastive loss is used as an objective function to learn the discriminative representations of the deep-learned features. The proposed framework is evaluated on three challenging datasets: COCO-2014, VOC2007 and VOC2012 datasets, where it shows superior performance compared with other approaches.  

The contribution of the proposed method is summarized as follows:
\begin{itemize}
    \item To the best of our knowledge, this is the first time a method is introduced to learn discriminative representations for the multi-label recognition problem. Quantitative and Qualitative results in Sec. \ref{sec:experiments} show the effectiveness of learning discriminative representations on improving the results of multi-label recognition task.
    \item A supervised object parser is proposed to disentangle the set of objects in an input image while training the model. This simplifies the task of learning discriminative representations of the object classes.   
    \item A contrastive loss function is proposed for the multi-label recognition task, which decreases intra-classes variations and increases inter-classes discrepancies.
    \item The proposed method is evaluated on COCO 2014, Pascal VOC 2007 and VOC 2012  datasets and showed significant improvement over the state-of-the art techniques.
\end{itemize}
\section{Related Work}
Multi-label recognition has been extensively explored in various ways such as learning separated label and correlated labels using multiple techniques such as RNN, graphical inferences, and attention networks \cite{srn,ml-gcn,struct_1}. Firstly, separated-labels' methods, e.g. Binary Relevance \cite{BR},  use normal classifiers to learn each class individually to check whether it is present in the output space or no. 
In \cite{HCPa}, Wei \textit{et al.} accumulates the outcomes of the object hypothesis using max pooling to get the final prediction of the multiple labels. In \cite{ml-ranking}, Gong \textit{et al.} rank the convolutional results to get the highest top-k representations. Though these methods achieved an improvement in the literature, they ignore the semantic relations between objects and labels that happen frequently in the real world.
%
Since correlated labels appear frequently in the same scene, a set of methods benefited multi-label recognition by encoding such relationships. CNN-RNN \cite{cnn-rnn} introduced a generic network with two subsequent modules: CNN to extract the objects' features, and RNN to encode the semantic relations of these objects. 
Similarly, graph-based methods have been used to encode the labels' correlations and proved their ability to improve performance. Li at el. \cite{li2016conditional} used the graphical lasso framework to extract the image-related conditional labels. In \cite{ml-zsl}, Lee \textit{et al.} modeled knowledge graphs to encode correlations between labels. Chen \textit{et al.} proposed using Graphical Convolutional Networks (GCN) to learn the correlations between the labels \cite{ml-gcn}. Although it represents the state-of-the art, it uses an external word embedding to supervise the training. 

Attention techniques are widely used in the visual recognition to force the network focus into the important features and capture the contextual information \cite{consis}. Zhe \textit{et al.} proposed a special module, namely Spatial Regularization Network (SRN), to encode the spatial correlation as well as semantic relations amongst objects using attention maps \cite{srn}. In \cite{consis}, Guo \textit{et al.} used attentions to enforce consistency for label-relevant regions under certain image transforms. Also, Wang \textit{et al.} used attention to extract the salient regions and used LSTM to encode these relations amongst labels \cite{struct_1}.

Learning discriminative features have been explored in several fields of visual recognition such as visual tracking \cite{discr_similar}, face recognition and verification \cite{centerloss}. Hu \textit{et al.} \cite{hu2014discriminative} trained nonlinear transformations to push the system to discriminate between the features by a margin between positive and negative pairs. Triplet loss was proposed to learn discriminative features for face recognition as they showed effectiveness in \cite{triplet2}. They used deep embedding to minimize the distance between an anchor and a positive anchor while it maximizes the distance towards the negative anchor. Overall, discrminaive learning in single-label recognition improved the accuracy and boosted the performance. However, it has not been explored in multi-label recognition because of the combinatorial nature of input space, latent space and output space.
\begin{figure*}[t]
    \centering
       \includegraphics[width=.8\textwidth]{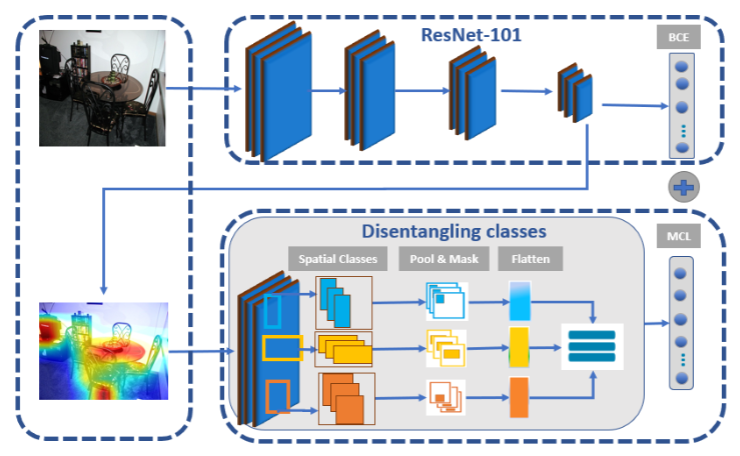}
   \caption{A visual representation of the training process of the proposed system. On the top of the figure, ResNet-101 is used as a backbone. The features are extracted from the penultimate layer and forwarded to the proposed loss function. The bottom row explains the processing steps to discriminate the features. These steps are as follows: extracting the class attention maps, obtaining object specific features via box localization and pooling, 
   and lastly ensuring class discrimination via minimizing intra-class distances and maximising inter-class distances.}
  \label{fig:system} 
  
\end{figure*}
\section{Proposed Method}
In this section, we describe the proposed approach to learn discriminative features for multi-label classification. The detailed architecture of the network is described in Figure \ref{fig:system}. It is built upon ResNet-101 \cite{Resnet_2016} as a representation learning backbone. 
The proposed loss operates on features from the penultimate layer of the network and discriminates them using the class-specific attention maps. It consists of two main steps: first breaking the many-to-many relation between features and classes into one-to-one. This is achieved by an object parser that converts the multi-label image to multiple object regions with only a single dominant label each. In the subsequent stage, a discriminative classifier is applied on the predicted features obtained from the object parser.

Suppose $I \in \mathbb{R}^{h \times  w \times 3}$ is an input color image with height $h$, width $w$ and an associated label set $Y \ =\  \{y_1,\ y_2,\ \ldots,\ y_n \}$ where $n$ is the total number of labels assigned to ${I}$. Here, $y_i \in \{1, \ldots,c\}$, where $c$ is the total number of categories in a given dataset. We consider a deep neural network model consisting of a feature extractor $f(\cdot)$ whose output is denoted by ${X} \in \mathbb{R}^{2048 \times 14 \times 14}$. The function $f(\cdot)$ generally comprises of a regular neural network from initial layers to the penultimate layer, excluding the final classification layer $h(\cdot)$. The features ${X}$ are then input to the object parser $g(\cdot)$ to generate bottom-up features that are subsequently sent to the proposed loss function $\mathcal{L}$ beside the ground-truth labels ${Y}$ to learn discriminative features. The whole process can be described as follows,
\begin{equation}
\begin{split}
{X}\ =\ f({I};\theta), \quad   \widehat{{Y}} = h({X}; \phi), \quad   {Z} = g({X}; \psi) \quad  \text{s.t., } \widehat{{Y}}  \in {\Bbb R} ^n
\end{split}
\end{equation}
\begin{equation}
\{\theta^*, \phi^*, \psi^*\} =  \underset{\theta, \phi, \psi}{\arg \min} (\lce + \lmc)
\end{equation}
where $\widehat{{Y}}$  is the set of predicted labels and $\mathcal{L}$ denote the loss functions.
%

Below, the object parser is firstly described 
in Section \ref{RB} to obtain category-specific features. Then, the discrimination between the obtained features using the proposed loss is elaborated in Section \ref{CCL}.
\subsection{Disentangling Class-specific Representations}\label{RB}
The multi-label problem considers that several labels exist for a single image. Given a feature extractor $f(\cdot)$ generating grid-level features ${X}$, a conventional model directly maps the features to $c$ object classes. We note that achieving discrimination in such a many-to-many mapping is fairly challenging, since the model considers all features to predict each class. In this case, the features from negative classes can interfere in the discrimination process. Therefore, we first simplify the task to a one-to-one mapping by using our object parser $g(\cdot)$ that separates image features into class-specific representations. 
%
 
  Specifically, we use the ground-truth bounding-box set $\mathcal{B} = \{{b}_1, \ldots, {b}_m\}$ with their corresponding class labels ${Y}_{\mathcal{B}} = \{\ell_1, \ldots,\ell_m\}$ to find class-specific features. Here, ${b}_i = \{x_i,y_i,h_i,w_i\} \in \mathcal{B}$, where $(x_i,y_i)$ are the top-left box coordinates and $h_i,w_i$ denote the height and width of $i$th bounding box in the image coordinates. Using these box coordinates, we obtain the normalized coordinates in the feature map space: ${b}'_i = \{x_i',y_i',h_i',w_i'\}$. For example, normalized box height and width is $h_i' = \nint{\frac{h_i}{h} \times 14}$ and $ w_i' = \nint{\frac{w_i}{w} \times 14}$, respectively. From this normalized set of bounding boxes, we identify boxes that belong to the same class ($j$) to create  category-specific sets of boxes, $\mathcal{B}'_j = \{{b}'_i : \ell_i = j \}$.
  
  Given the set $\mathcal{B}'_j$ for each class present in the image, we create a class-wise spatial mask ${M}_j \in \{0,1\}^{14\times 14}$ whose elements are zero only at the spatial locations of negative classes (\textit{i.e.}, $\{k \neq j\}$). This can be represented as, 

\begin{figure}
       \includegraphics[width=\textwidth]{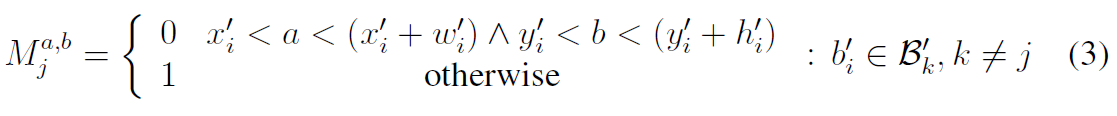}
\end{figure}
 where, $a,b$ are the matrix indices such that ${M}^{a,b}_j \in {M}_j $. Note that the mask will be `On' at all the bounding boxes of a class $j$ as well as the background regions. This is an important design choice driven by our observation that the inclusion of background allows the model to incorporate image context  which helps to learn enriched representations. The final class specific features are then obtained by applying the class-specific mask and pooling the features along spatial dimensions. This step is given as follows,
\begin{align}
     {Z} = \{{z}_j\} \text{  where,} \quad {z}_j = \text{maxpool}({X} \circ {M}_j), \quad  j \in {Y}, {z}_j \in \mathbb{R}^{2048}.
     \label{Eq:pooling}
 \end{align}

Since the given image ${I}$ has a total of $n$ labels, the set cardinality of ${Z}$ is also $n$.
\subsection{Multi-label Contrastive Loss}
\label{CCL}
In this section, we introduce our Multi-label Contrastive (MC) loss that encourages the network to learn discriminative features. MC loss is based on the Center loss \cite{centerloss}, however for the first time, we demonstrate how it can be tailored for multi-label recognition problems to enhance class discrimination. Notably, the center loss can not be directly used for multi-label learning since it considers a direct relationship between image features and class centers. Since, in multi-label learning, a single feature can relate to multiple class centers, we first disentangle the image features into class-specific features as described in Sec.~\ref{RB}.
This leads to a straight-forward definition of Center Loss (CL) in our context that seeks to minimize the distance between features ${z}_j$ belonging to class $j$ and the learned center ${c}_{j}$ of the true class,
\begin{equation}
\mathcal{L}_{\text{CL}} = \frac{1}{2n} \sum_{i=1}^{r}\sum_{j=1}^n{\parallel {z}^i_j - {c}_j \parallel_{2}^2},
\label{Eq:l1}
\end{equation}
where $r$ denotes the batch size.
However, the above loss function does not consider contrastive constraints between classes, e.g., the distances between difference classes are not explicitly maximized. To this end, we propose MC loss function ($\lmc$) that seeks to reduce inter-class distances (first term in $\lmc$) and simultaneously increases inter-class discrimination (second term in $\lmc$), as follows:
\begin{align}
   \lmc = \frac{1}{2n}\sum_{i=1}^{r}\sum_{j=1}^n\Big({\parallel {z}^i_j - {c}_j \parallel_{2}^2} - \beta \sum_{k \neq j}{\parallel {z}^i_k - {c}_j \parallel_{2}^2} \Big), 
   \label{Eq:lmc}
\end{align}
where, ${z}_k$ denote the features from the same $i$th image in the batch, that do not correspond to the class $j$ (\textit{i.e.}, $k \neq j$). The factor $\beta$ is used weight the trade-off between contrastive terms in the loss function. 


The class centers $\{{c}_j\in \mathbb{R}^{2048}\}$ are learned in order to achieve better discriminability. Following \cite{centerloss}, centers are updated in each training iteration as: 
\begin{align}
    \Delta {c}_j = \frac{\sum_{i=1}^r \sum_{j=1}^{c} [\![ j \in {Y}^i ]\!] ({c}_j - {z}^i_j)}{ 1 + \sum_{j=1}^{c} [\![ j \in {Y}^i ]\!] }, 
    \label{Eq:der}
\end{align}
where $[\![\cdot ]\!]$ denote the Iverson brackets that evaluate to 1 if the condition is true. 

The function $h(\cdot)$ in the top branch of Figure \ref{fig:system} operates on ${X}$ and generates logit scores ${p} = h({X}) \in \mathbb{R}^{c}$. The result is then fed to sigmoid modules $\sigma(p_i) = \frac{1}{1 +  e^{-p_i}}$ to get the values between $0$ and $1$. Then, the binary Cross Entropy loss $\lce$ is applied, which takes into account the implicit discrepancy in prediction for each class compared to the ground-truth,
\begin{align}
\lce = -\sum_{i} [\![ i \in {Y}]\!] \ \log\ \sigma(p_i) +  [\![ i \not\in {Y}]\!] \ \log(1-\ \sigma(p_i)).
\label{Eq:ce}
\end{align}

The overall framework is trained with joint supervision, where the total objective function is the summation of binary cross entropy (Eq.~\ref{Eq:ce}) and multi-label contrastive loss (Eq.~\ref{Eq:lmc}), as follows:
\begin{align}
\mathcal{L} = \lce + \lmc
\label{Eq:total_loss}
\end{align}

This ensures the proposed framework is trained to extract discriminative class-specific representations, where inter-class discrepancies are maximised and intra-class variations are minimised. Algorithm \ref{algorithm} summarizes the training procedures of the proposed framework.
%
%
\begin{algorithm}[t]
\SetAlgoLined
\SetKwInOut{Input}{Input}
\SetKwInOut{Output}{Output}
 \Input {Training data $\{I\}$, Initialized parameters $\theta_o, \phi_o, \psi_o$, initial weights $w_o$ of class centers $\{c_j | j= 1,2, ..., n\}$, $epochs$ for maximum number of iterations and ${b}_i = \{x_i,y_i,h_i,w_i\} \in \mathcal{B}$, where $\mathcal{B}$ is set of bounding boxes}
 \Output{Set of trained parameters $\{\theta^*, \phi^*, \psi^*\}$}
 $t = 0$
    \While{$t < epochs$}{
    $t = t+ 1$
    \While{$b_i < n$}{
     Disentangling class-specific features (Eq. \ref{eq:mask})
     Max pooling these features (Eq. \ref{Eq:pooling})
    }
     Compute the contrastive loss $\lmc$ (Eq. \ref{Eq:lmc})
     Compute the joint loss $\mathcal{L} = \lce + \lmc$ (Eq. \ref{Eq:total_loss}) 
     Compute gradients $\bigtriangledown_{\theta, \phi, \psi}\mathcal{L}$ 
     Update centers' weights $c\ \forall c_j$, (Eq. \ref{Eq:der})
     Update the set of parameters ${\theta, \phi, \psi}$}

 return $\{\theta^*, \phi^*, \psi^*\}$
 \caption{Discriminative learning for multi-label classification}
 \label{algorithm}
\end{algorithm}
%
\section{Experiments}
\label{sec:experiments}
In this section, we present implementation details of the proposed method and evaluation metrics on three datasets: MS-COCO 2014 \cite{coco}, Pascal VOC 2007 and 2012 \cite{pascal}. Then, the comparison with the state-of-the art is summarized quantitatively and qualitatively. This is followed by the ablation analysis to evaluate the impact of different components on the performance of the proposed method.
\subsection{Implementation Details}
Without otherwise stated, the proposed method has the same settings as of ResNet-101 \cite{Resnet_2016}. Regarding discriminative settings, the shape of the penultimate layer is  $2048 \times 14 \times 14$, for channel size, width and height respectively. The length of the mask (mentioned in Eq. 3 ) is selected to be $2048$ to align with the feature size. The kernel of the max pooling is $14$. The trainable weights, $c_j$, are initialized as a vector of size $r \times 2048$, where $r$ is the number of classes. The trainable weights are optimized using SGD \cite{sgd} with a learning rate $0.5$. However, for training the whole network, SGD is selected as the optimizer with a learning rate, which starts from $0.01$ and decays by a rate of $10$ after every $30$ epochs. The selected momentum is set to 0.9 whereas the weight decay is $10^{-4}$. The network parameters are initialized by the values, which have been pre-trained on the ImageNet dataset \cite{Imagenet_2015}. Eventually, the total number of epochs for training the proposed model is $100$.
\begin{table*}[t]
\scalebox{1}{
\centering
\resizebox{\linewidth}{!}{\begin{tabular}{|c|ccccccc|cccccc|}
\hline 
{Method} &\multicolumn{7}{|c|}{\textbf{All}}&\multicolumn{6}{|c|}{\textbf{Top-3}}\tabularnewline
&mAP &CP&CR&CF1&OP&OR&OF1&CP&CR&CF1&OP&OR&OF1 \tabularnewline
\hline
CNN-RNN \cite{cnn-rnn} &61.2 & -  & -  & -  & - & - & -  & 66.0 & 55.6 & 60.4 & 69.2 & 66.4 &67.8 \\
RNN-Attention \cite{struct_1} & - & -& -&- &- &-&- &79.1 &58.7 &67.4 &84.0 &63.0 &72.0\\
Order-Free RNN \cite{orderfree} & - & -& -&- &- &-&- &71.6 &54.8 &62.1 &74.2 &62.2 &67.7\\
ML-ZSL \cite{ml-zsl} & - & -& -&- &- &-&- &74.1 &64.5 &69.0 &- &- &-\\
SRN \cite{srn} & 77.1& 81.6& 65.4&71.2 &82.7 &69.9&75.8 &85.2 &58.8 &67.4 &87.4 &62.5 &72.9\\
ResNet-101 \cite{ml-gcn}  & 77.3& 80.2& 66.7&72.8 &83.9 &70.8&76.8 &84.1 &59.4 &69.7 &89.1 &62.8 &73.6\\
Multi-Evidence \cite{evidence}  & -& 80.4&\textbf{ 70.2}&74.9 &85.2 &72.5&78.4 &84.5 &\textbf{62.2} &70.6 &89.1 &64.3 &74.7\\
Orderless \cite{2020orderless}  & -& 80.38&68.85&74.17 &81.46&72.26&77.14 & -&-&-&-&-&-\\
Visual Consistency \cite{consis}  & 77.5& 77.4&68.3&72.2 &79.8 &73.1&76.3 &85.2 &59.4 &68.0 &86.6 &63.3 &73.1\\
\hline
Proposed method   &\textbf{ 81.0}& \textbf{82.7}& 69.5 & \textbf{75.6} &\textbf{85.5}      &\textbf{73.4}      &\textbf{79.3} &\textbf{86.6} &62.1 &\textbf{72.3} &\textbf{90.0} &\textbf{65.4} &\textbf{75.8}\\
\hline
Relative Improvement   &\textbf{ 3.5\%}& \textbf{1.1\%}& \textbf{-0.7\%} & \textbf{0.7\%} &\textbf{0.3\%}      &\textbf{0.3\%}      &\textbf{1.1\%} &\textbf{0.9\%} &\textbf{-0.1\%} &\textbf{2.4\%} &\textbf{0.4\%} &\textbf{1.1\%} &\textbf{1.1\%}\\
\hline
\end{tabular}}
}
\newline
\caption{\small{Comparison between the proposed method with state-of-the-art techniques on the MS-COCO dataset.}}
\label{table:coco}
\end{table*}
\subsection{Evaluation Metrics}
To evaluate the quantitative performance of the proposed method, we have used the multi-label evaluations metrics as in \cite{metrics}. These metrics represent two groups as follows: 1) the per-class metrics (average per-cass precision (CP), recall (CR), F1-score (CF1)) and 2) the overall metrics (the average overall precision (OP), recall (OR), F1-score (OF1)). Also, the mean Average Precision (mAP), which represents the mean value of per-class average precision is used as an evaluation metric. These metrics can be evaluated with a flexible number of labels for every image. Generally, mAP, CF1 and OF1 are the most important metrics. The Top-3 labels evaluation approach is utilized to provide a fair comparison with the baseline techniques. Lastly, recall and F1 evaluation metrics follow the constraints of top-3 labels at every image.
\subsection{Quantitative Results}
\paragraph{MS-COCO:}
Microsoft COCO \cite{coco} is one of the most challenging datasets for visual recognition paradigms. It has $82,081$ images for training and $40,504$ images for validation. The number of classes in this dataset is $80$. The distribution of the labels varies dynamically, which makes it more challenging. 
The proposed method is compared against $9$ methods as follows: 1) SRN \cite{srn} 2) ResNet-101 \cite{ml-gcn} 3) CNN-RNN \cite{cnn-rnn} 4) Visual consistency \cite{consist} 5) RNN-Attention \cite{struct_1} 6) Order-Free RNN \cite{orderfree} 7) ML-ZSL \cite{ml-zsl} 8) Multi-Evidence \cite{evidence} 9) Visual consistency \cite{consis}.

The evaluation is performed on the validation set to provide fair comparison with the baseline approaches as the annotation of the test set is not publicly available. The quantitative results of the proposed method compared with the baseline approaches are summarized in Table \ref{table:coco}. These results present a strong evidence  that the proposed method outperform the baseline approaches in most of the evaluation metrics. More importantly, it surpasses the state-of-the-art by $3.5\%$ in mAP. Based on these results, the proposed method introduces a new state-of-the-art results for the multi-label recognition task on the MS-COCO dataset.

The proposed approach outperforms all the baselines by a reasonable margin, which varies from $0.3\% $ to $3.5\%$, except in one case (for the CR evaluation metric, which is less important metric). For, the most important metrics, (mAP, CF1 and OF1), the proposed method achieves $3.5\%$, $0.7\%$ and $1.1\%$ as a relative improvement, respectively. Similarly, it exceeds the baselines in the Top-3 settings for the most important metrics, CF1 and OF1, with $2.4\%$ and $1.1\%$, respectively. It is worth noting that the proposed method outperforms all the baselines in all the evaluation metrics except the Multi-Evidence \cite{evidence}, which reported higher outcomes in two metrics only (CR for All and Top-3 settings). However, CR is not one of the important metrics as mentioned earlier because it calculates the percentage between the correct labels and the total of the predicted labels.
\paragraph{VOC 2007:} PASCAL Visual Object Classes Challenge (VOC 2007) is another common dataset for visual recognition tasks \cite{PASCAL_DS}. It consists of $9,963$ images, where half of them are hold for training and validation purpose, and the other half are for testing purpose. The number of categories is $20$ classes. We selected the training and validation sets to train and validate the proposed model, respectively. Eight methods have been selected as baselines, where they are compared with proposed method. These baselines are as follows: 1) CNN-RNN \cite{cnn-rnn}, RSLD \cite{RLSD}, Very Deep \cite{verydeep}, ResNet-101 \cite{ml-gcn}, FeV+LV \cite{FLV}, HCP\cite{HCP}, RNN-Attention \cite{struct_1} and Atten-Reinforce \cite{recurrent_atte}. The Average Precision (AP) and the mean Average Precision (mAP) evaluation metrics are used to evaluate the proposed method and to provide fair comparison with the baselines. 

The comparison with the baselines is presented in Table \ref{table:voc}. The results illustrate that learning discrminative representations from the input images is able to outperform the baselines and introduces a new state-of-the art results. Furthermore, the proposed method reported higher performance in all of classes except $5$ categories only. While the most important metric in this evaluation is the mean average precision (mAP), and our approach provides a relative improvement of $1.4\%$ above the baselines approaches.
\begin{table*}[h]
\centering
\scalebox{1}{
\resizebox{\textwidth}{!}{\begin{tabular}{|l|c|c|c|c|c|c|c|c|c|c|c|c|c|c|c|c|c|c|c|c|c|}
\hline 
Method & aero & bike & bird & boat & bottle & bus & car & cat & chair & cow & table & dog & horse & motor & person & plant & sheep & sofa & train& tv & mAP\tabularnewline
\hline 
CNN-RNN \cite{cnn-rnn} &96.7& 83.1& 94.2& 92.8& 61.2& 82.1& 89.1& 94.2& 64.2& 83.6& 70.0& 92.4& 91.7& 84.2& 93.7& 59.8& 93.2& 75.3& \textbf{99.7}& 78.6& 84.0
\tabularnewline
RLSD \cite{RLSD}& 96.4& 92.7& 93.8& 94.1& 71.2& 92.5& 94.2& 95.7& 74.3& 90.0& 74.2& 95.4& 96.2& 92.1& 97.9& 66.9& 93.5& 73.7& 97.5& 87.6& 88.5\tabularnewline
VeryDeep \cite{verydeep}&98.9&95.0&96.8&95.4&69.7&90.4&93.5&96.0&74.2&86.6&87.8&96.0&96.3&93.1&97.2&70.0&92.1&80.3&98.1&87.0&89.7\tabularnewline
ResNet-101 \cite{ml-gcn}&99.5&97.7&97.8&96.4&65.7&91.8&96.1&97.6&74.2&80.9&85.0&98.4&96.5&95.9&\textbf{98.4}&70.1&88.3&80.2&98.9&89.2&89.9\tabularnewline
FeV+LV \cite{FLV}& 97.9&97.0&96.6&94.6&73.6&93.9&96.5&95.5&73.7&90.3&82.8&95.4&\textbf{97.7}&95.9&98.6&77.6&88.7&78.0&98.3&89.0&90.6\tabularnewline
HCP\cite{HCP}&98.6&97.1&98.0&95.6&75.3&\textbf{94.7}&95.8&97.3&73.1&90.2&80.0&97.3&96.1&94.9&96.3&78.3&94.7&76.2&97.9&91.5&90.9\tabularnewline
RNN-Attention \cite{struct_1}&98.6&97.4&96.3&96.2&75.2&92.4&96.5&97.1&76.5&92.0&\textbf{87.7}&96.8&97.5&93.8&98.5&81.6&93.7&82.8&98.6&89.3&91.9\tabularnewline
Atten-Reinforce \cite{recurrent_atte} &98.6&97.1&97.1&95.5&75.6&92.8&96.8&97.3&78.3&92.2&87.6&96.9&96.5&93.6&98.5&81.6&93.1&83.2&98.5&89.3&92.0\tabularnewline
\hline
Proposed Method & \textbf{99.6}& \textbf{98.3}& \textbf{98.0}& \textbf{98.2}& \textbf{78.2}& 94.2& \textbf{97.0}& \textbf{97.8}&
        \textbf{80.8}& \textbf{94.9}& 84.9& 97.7& 97.5& \textbf{96.6}& \textbf{98.7}& \textbf{85.0}&
        \textbf{96.2}& \textbf{83.2}& 98.5& \textbf{92.6} & \textbf{93.4}
 \tabularnewline
\hline
Relative Improvement &\textbf{0.1\%}& \textbf{0.6\%}& \textbf{0.0\%}& \textbf{1.8\%}& \textbf{2.6\%}& \textbf{-0.5\%}& \textbf{0.2\%}& \textbf{0.2\%}&
        \textbf{2.5\%}& \textbf{2.7\%}& \textbf{-2.8\%}& \textbf{-0.7\%}& \textbf{-0.2\%}& \textbf{0.7\%}& \textbf{0.2\%}& \textbf{3.5\%}&
        \textbf{1.5\%}& \textbf{0\%}& \textbf{-1.2\%}& \textbf{1.1\%} & \textbf{1.4\%}
 \tabularnewline
\hline
\end{tabular}}
}
\caption{\small{Comparison between the proposed method and state-of-the-art methods on the VOC2007 dataset \cite{PASCAL_DS}.}} \label{table:voc}
\end{table*}
\paragraph{VOC 2012 PASCAL:} VOC 2012 is another version of Pascal VOC dataset with a slight difference in the number of images of train-validation and test sets. It consists of $11,540$ images for train-validation set and $10,991$ for test set, where the number of categories is the same as VOC 2007 (\textit{i.e.} $20$ classes) \cite{PASCAL_DS}. Because there are less experiments in the literature on VOC 2012, we compare the proposed method with $6$ baseline techniques only. These baselines are as follows: 1) RMIC \cite{he2018reinforced}, 2) VGG16+SVM \cite{verydeep}, 3) VGG19+SVM \cite{verydeep}, 4) HCP\cite{HCP}, 5) FeV+LV \cite{FLV}, 6) RCP \cite{rcp}. The Average Precision (AP) and mean Average Precision (mAP) evaluation metrics are used to evaluate the proposed method against the baselines.
\begin{table*}[]
\scalebox{1}{
\centering
\resizebox{\textwidth}{!}{\begin{tabular}{|l|c|c|c|c|c|c|c|c|c|c|c|c|c|c|c|c|c|c|c|c|c|}
\hline 
Method & aero & bike & bird & boat & bottle & bus & car & cat & chair & cow & table & dog & horse & motor & person & plant & sheep & sofa & train& tv & mAP\tabularnewline
\hline 
RMIC \cite{he2018reinforced} &98.0&85.5&92.6&88.7&64.0&86.8&82.0&94.9&72.7&83.1&73.4&95.2&91.7&90.8&95.5&58.3&87.6&70.6&93.8&83.0&84.4
\tabularnewline
 VGG16+SVM \cite{verydeep}&99.0&88.8&95.9&93.8&73.1&92.1&85.1&97.8&79.5&91.1&83.3&97.2&96.3&94.5&96.9&63.1&93.4&75.0&97.1&87.1& 89.0\tabularnewline
VGG19+SVM \cite{verydeep}&99.1&88.7&95.7&93.9&73.1&92.1&84.8&97.7&79.1&90.7&83.2&97.3&96.2&94.3&96.9&63.4&93.2&74.6&97.3&87.9&89.0\tabularnewline
HCP\cite{HCP}&99.1&92.8&97.4&94.4&79.9&93.6&89.8&98.2&78.2&94.9&79.8&97.8&97.0&93.8&96.4&74.3&94.7&71.9&96.7&88.6&90.5\tabularnewline
FeV+LV \cite{FLV}&98.4&92.8&93.4&90.7&74.9&93.2&90.2&96.1&78.2&89.8&80.6&95.7&96.1&95.3&97.5&73.1&91.2&75.4&97.0&88.2&89.4\tabularnewline
RCP \cite{rcp} &99.3&92.2&97.5&94.9&82.3&94.1&92.4&98.5&83.8&93.5&83.1&98.1&97.3&96.0&98.8&77.7&95.1&79.4&97.7&92.4&92.2\tabularnewline
SSGRL \cite{SSGRL} &99.5&95.1&97.4&96.4&85.8&94.5&93.7&98.9&86.7&96.3&84.6&98.9&98.6&96.2&98.7&82.2&98.2&\textbf{84.2}&98.1&93.5&93.9\tabularnewline
\hline
MCL &\textbf{99.5}& \textbf{96.6}& \textbf{98.7}& \textbf{96.6}& \textbf{86.4}& \textbf{96.5}& \textbf{94.2}& \textbf{99.6}&
        \textbf{87.5}& \textbf{97.7}& \textbf{87.1}& \textbf{99.5}& \textbf{98.9}& \textbf{97.4}& \textbf{98.9}& \textbf{82.2}&
        \textbf{98.7}& 82.7& \textbf{99.4}& \textbf{94.9} & \textbf{94.7}
 \tabularnewline
\hline
Relaive Improvement &\textbf{0.0\%}& \textbf{1.5\%}& \textbf{1.2\%}& \textbf{0.2\%}& \textbf{0.6\%}& \textbf{2.0\%}& \textbf{0.5\%}& \textbf{0.7\%}&
        \textbf{0.8\%}& \textbf{1.4\%}& \textbf{2.5\%}& \textbf{0.6\%}& \textbf{0.3\%}& \textbf{1.2\%}& \textbf{0.2\%}& \textbf{0.0\%}&
        \textbf{0.5\%}& \textbf{-1.5\%}& \textbf{1.3\%}& \textbf{1.4\%} & \textbf{0.8\%}
 \tabularnewline
\hline
\end{tabular}}
}
\newline
\caption{\small{Comparison between the proposed method and state-of-the-art techniques on the VOC2012 dataset.}} 
\label{table:voc12}
\end{table*}

The results in Table \ref{table:voc12} illustrate that the proposed method outperforms the other baseline techniques and even for the complementary-guided methods, which employ word embedding to supervise the training process, such as in \cite{SSGRL}. the proposed method has achieved the highest mAP $94.7\%$, where the closest approach \cite{SSGRL} achieved $93.9$. More specifically, these results reveal that the proposed method has achieved better performance in $17$ out of the $20$ classes, while it has been less than the baselines for only one class (\textit{i.e.} sofa).

\begin{figure}[h]{
      \includegraphics[width=\textwidth]{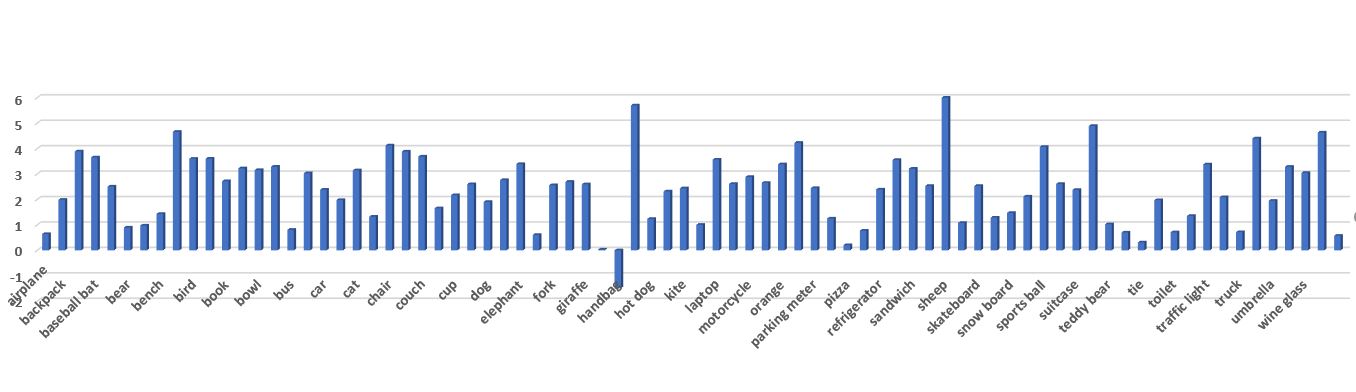}}
  \caption{The improvement of MCL for all the COCO classes.}
  \label{fig:improvements} 
\end{figure}

Many of the baseline techniques have encoded the correlation between labels of the objects while training as in \cite{ml-gcn,SSGRL,cnn-rnn}. Learning this correlation from textual data is much easier than from visual data because of challenges such as occlusion, illumination changes, unbalanced datasets, etc. exist in visual data. Recently, \cite{ml-gcn}  and \cite{SSGRL} have used a ready-made correlation matrix trained on Wikipedia dataset for COCO labels to provide a self-guidance for the correlation between the labels. Because of this complementary supervision, \cite{ml-gcn} has achieved an mAP of $80.3\%$ for automatic settings and $83.0\%$ for manual settings on MS-COCO dataset. Specifically, they have minimized the participation of negative classes on the scene in order to maximize the correlations between the positive classes. Also, \cite{SSGRL} used a similar technique (\textit{i.e.} complementary supervision), however it was processed with Long-Short Term Memory (LSTM) rather than using Graphical Convolutional Network (GCN). Although these methods have achieved significant improvement with the complementary supervision, our proposed method has been able to achieve comparable results with their techniques without any complementary supervision, further architectures (GCN, LSTM) nor extra fusion layers. In \ref{fig:improvements}, the improvements for COCO classes is displayed. MCL regularization benefits the unbalance in the datasets. For example, long-tail classes (tooth brush, scissors, bear, parking meter, toaster and hair drier) and small object classes (handbag, mouse, bottle, baseball glove, traffic glove and sport ball). From the figure, it is clear that MCL improves the balance for unbalanced datasets. For instance, baseball glove class in the number 36 reports over $5.5$ improvement. From small object cases, hair drier reports $\approx 5$ before the last class in the figure.
%
\subsection{Qualitative Results}
In this section, we present the significance of the proposed method qualitatively. We observe that learning discriminative representations helps multi-label recognition in the essence that it pulls same classes' features towards their centers and pushes away the features of other classes.

\begin{figure*}[h]
  \centering
  \texttt{The proposed approach}
  
  \hrulefill
  
  \medskip
  \begin{subfigure}[t]{.24\linewidth}
    \centering\includegraphics[width=\linewidth]{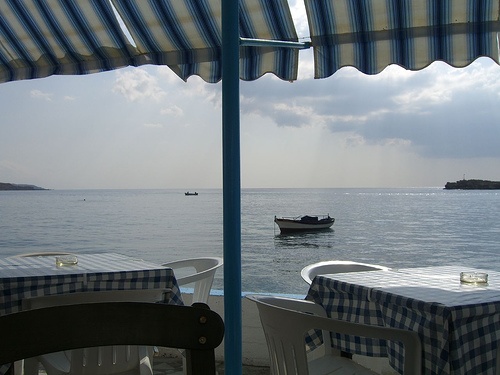}
    \caption{ \centering boat,chair,dintable  \green{boat, chair,} \red{person}}
  \end{subfigure}
\begin{subfigure}[t]{.24\linewidth}
    \centering\includegraphics[width=\linewidth]{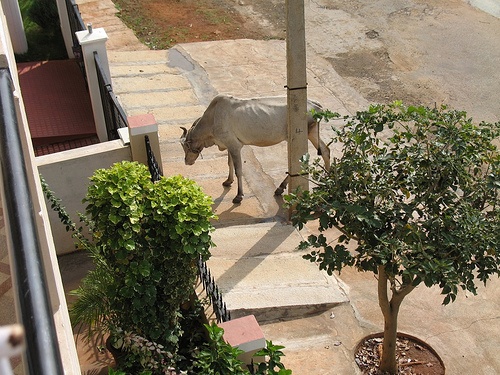}
    \caption{ \centering cow,pot,plant \green{cow}, \red{horse}}
  \end{subfigure}
  \begin{subfigure}[t]{.24\linewidth}
    \centering\includegraphics[width=\linewidth]{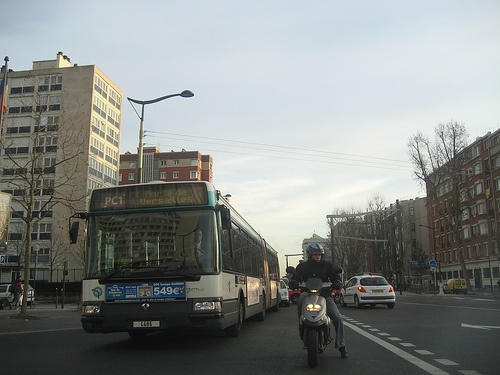}
    \caption{\centering bus,car,mobike,person \green{car,bus,person,mobike}}
  \end{subfigure}
   \begin{subfigure}[t]{.24\linewidth}
    \centering\includegraphics[width=\linewidth, height=.0907\paperheight]{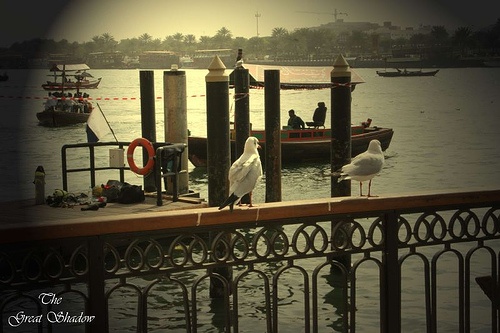}
    \caption{\centering bird,boat \green{bird}, \red{person}}
  \end{subfigure}

  \texttt{ResNet-101}
  
  \hrulefill
  
  \begin{subfigure}[t]{.24\linewidth}
    \centering\includegraphics[width=\linewidth]{figures/002449.jpg}
    \caption{ \centering boat,chair,dintable \green{ dintable},\red{sofa,mobike}}
  \end{subfigure}
\begin{subfigure}[t]{.24\linewidth}
    \centering\includegraphics[width=\linewidth]{figures/004081.jpg}
    \caption{\centering cow,pot,plant  \red{train, dog}}
  \end{subfigure}
  \begin{subfigure}[t]{.24\linewidth}
    \centering\includegraphics[width=\linewidth]{figures/004305.jpg}
    \caption{ \centering bus,car,mobike,person \green{bus,mobike},\red{train,dintable}}
  \end{subfigure}
   \begin{subfigure}[t]{.24\linewidth}
    \centering\includegraphics[width=\linewidth, height=.0907\paperheight]{figures/004443.jpg}
    \caption{\centering bird,boat \red{train,chair}}
  \end{subfigure}
\caption{Comparison of predictions of the proposed method and vanilla ResNet-101 \cite{Resnet_2016}. Top row shows the proposed method predictions while the output of ResNet-101 is shown in the bottom row. At each row, top labels are the ground-truth ones while the models' predictions are given below. Green values indicate correct predictions, while red ones represent errors in predictions.}
  \label{fig:labels_compare}
\end{figure*}

In Figure \ref{fig:labels_compare}, four images from the VOC 2007 dataset are shown in order to compare between the proposed method and the a baseline technique (\textit{i.e. } vanila ResNet-101 \cite{Resnet_2016}). The proposed method was able to predict most of the labels correctly as shown in the top part of the figure, while ResNet-101 exhibits issues in predicting correctly, particularly for the small objects.

\begin{figure*}[h]
\centering
Image \quad\quad\quad Ours \quad\quad \quad ResNet \quad\quad\quad Image \quad \quad \quad Ours\quad  \quad\quad ResNet

  \hrulefill

  \medskip

        \begin{subfigure}[b]{0.48\textwidth}
    \includegraphics[width=0.28\linewidth, height=0.05\paperheight]{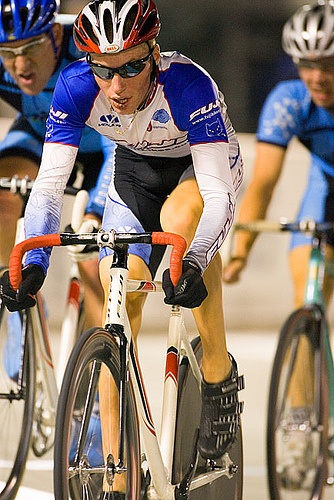}
       \hfill
         \includegraphics[width=0.28\linewidth, height=0.05\paperheight]{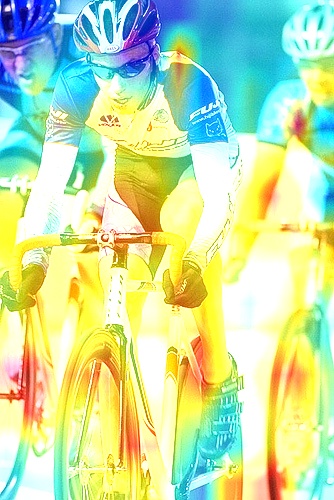}
       \hfill
         \includegraphics[width=0.28\linewidth, height=0.05\paperheight]{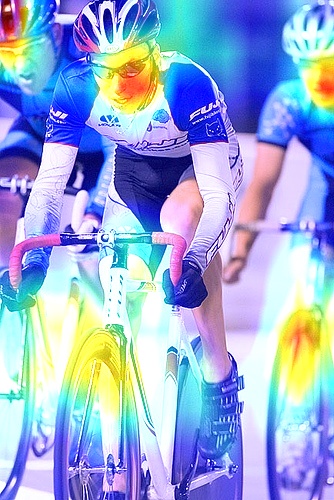}
       \hfill
        \includegraphics[width=0.28\linewidth, height=0.05\paperheight]{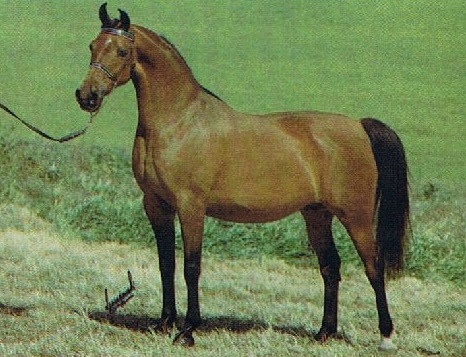}
       \hfill
        \includegraphics[width=0.28\linewidth, height=0.05\paperheight]{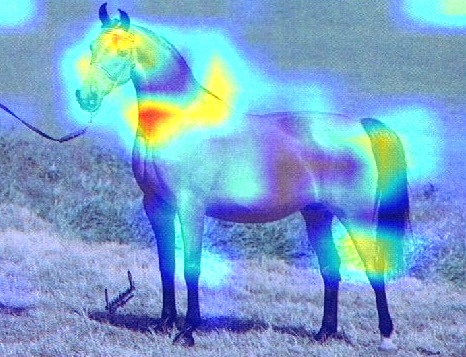}
       \hfill
        \includegraphics[width=0.28\linewidth, height=0.05\paperheight]{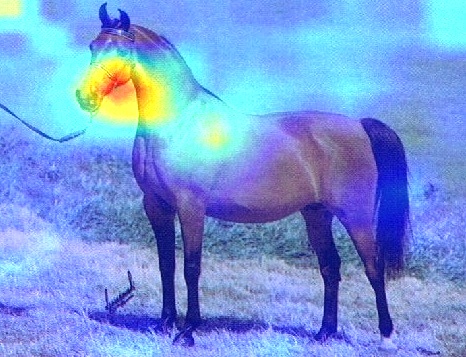}
        \hfill
        \includegraphics[width=0.28\linewidth, height=0.05\paperheight]{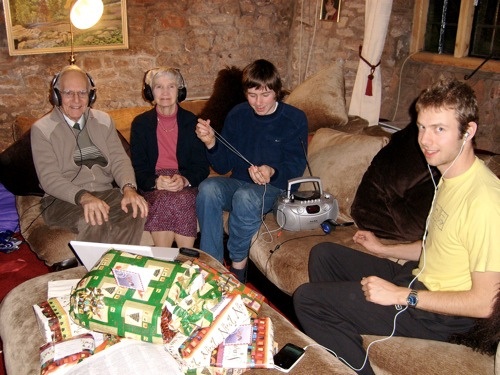}
       \hfill
         \includegraphics[width=0.28\linewidth, height=0.05\paperheight]{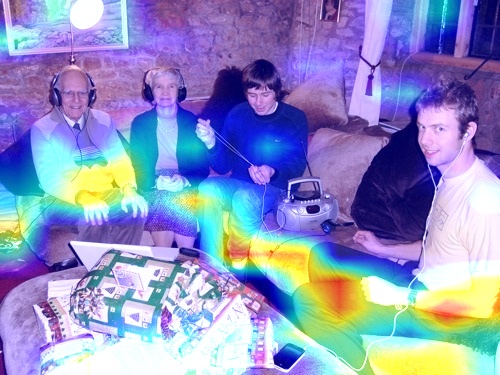}
       \hfill
         \includegraphics[width=0.28\linewidth, height=0.05\paperheight]{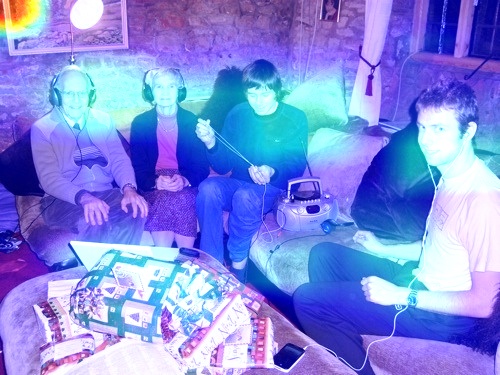}
    \caption{}
     \end{subfigure}
    \hskip\baselineskip
    \begin{subfigure}[b]{0.46\textwidth}
        \includegraphics[width=0.28\linewidth, height=0.05\paperheight]{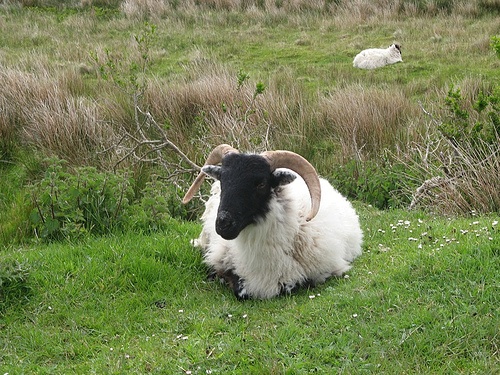}
       \hfill
        \includegraphics[width=0.28\linewidth, height=0.05\paperheight]{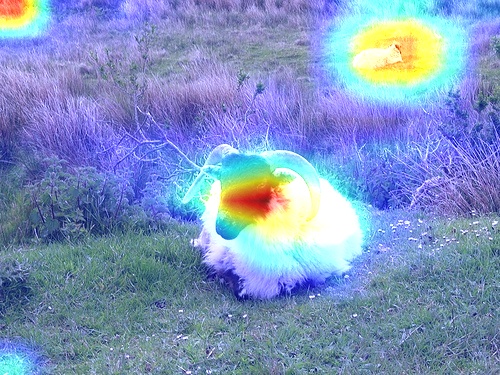}
        \hfill
        \includegraphics[width=0.28\linewidth, height=0.05\paperheight]{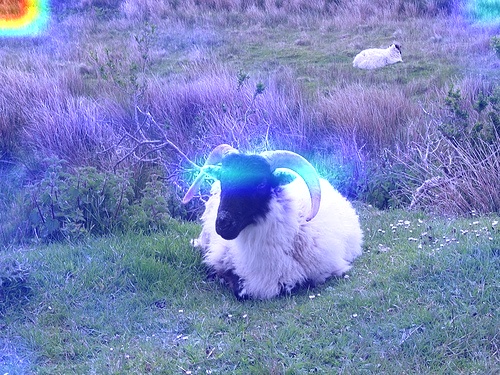}
        \hfill
        \includegraphics[width=0.28\linewidth, height=0.05\paperheight]{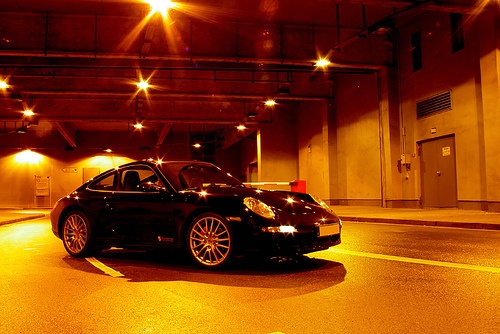}
        \hfill
         \includegraphics[width=0.28\linewidth, height=0.05\paperheight]{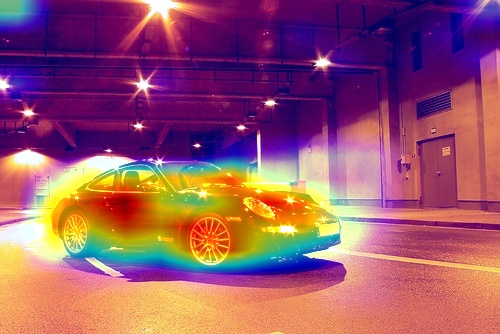}
        \hfill
         \includegraphics[width=0.28\linewidth, height=0.05\paperheight]{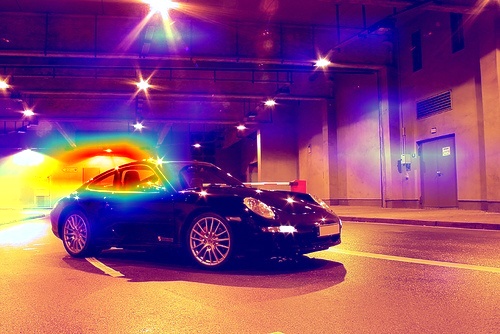}
         \hfill
         \includegraphics[width=0.28\linewidth, height=0.05\paperheight]{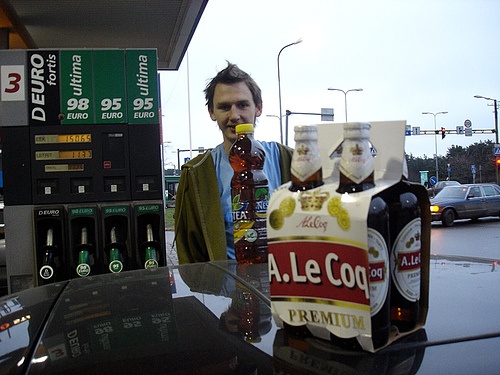}
       \hfill
         \includegraphics[width=0.28\linewidth, height=0.05\paperheight]{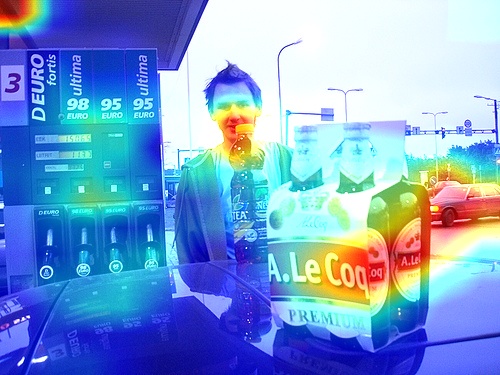}
       \hfill
         \includegraphics[width=0.28\linewidth, height=0.05\paperheight]{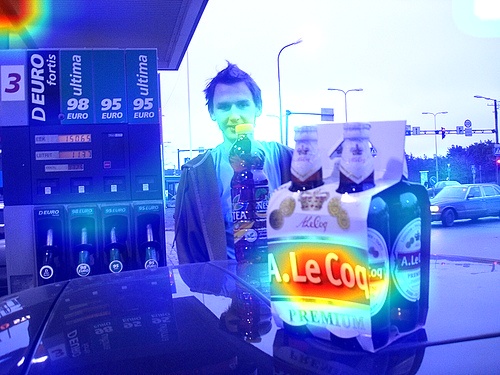}
       \hfill
        \caption{}
     \end{subfigure}
    \caption{A visual comparison of the attention maps for example from VOC dataset between the proposed method and ResNet-101 \cite{Resnet_2016} . The first column shows original images, second row shows the learned attention maps produced by our method and the third column are the maps produced using ResNet-101.}
    \label{fig:attentiones}
\end{figure*}

In Figure \ref{fig:attentiones}, the attention maps of the proposed method and ResNet-101 are shown in the following order: original images, attention maps of the proposed method and attention maps of using vanilla ResNet-101. The visual comparison ensures the importance of learning discriminative representations for the multi-label recognition, because of the high confusion of multiple instances that share the same feature space in the single batch. The maps predicted from the proposed method are more attending within the instances, where they are less attending in the gaps between the instances.
%



\subsection{Ablation Study}
As shown above, learning discriminative features results in performance improvement in three datasets. In this section, we focus on demonstrating the power of the proposed method. Firstly, an experiment for multi-class softmax over classes' attention heads is compared with MCL. Secondly, the effectiveness of various components of the proposed Multi-label Contrastive Loss, mentioned in Section \ref{CCL}. This includes the different values of the contrastive-hyper parameter $\beta$, first and second parts of Eq. \ref{Eq:lmc}. The first term of Eq. \ref{Eq:lmc} represents the intra-classes minimization component, while the second part represents the inter-classes maximization component.
\paragraph{\textbf{Comparison with multi-class softmax}}
The multi-class cross entropy is fixed on top of the attention heads that output of disentangling  the features as shown in Eq. \ref{eq:mask} and \ref{Eq:pooling}.

\begin{table*}[h]
\centering
\scalebox{1}{
\resizebox{\textwidth}{!}{\begin{tabular}{|l|c|c|c|c|c|c|c|c|c|c|c|c|c|c|c|c|c|c|c|c|c|}
\hline 
Method & aero & bike & bird & boat & bottle & bus & car & cat & chair & cow & table & dog & horse & motor & person & plant & sheep & sofa & train& tv & mAP\tabularnewline

     \hline
     
     BCE \cite{ml-gcn}&99.5&97.7&97.8&96.4&65.7&91.8&96.1&97.6&74.2&80.9&85.0&98.4&96.5&95.9&\textbf{98.4}&70.1&88.3&80.2&98.9&89.2&89.9\tabularnewline
CE & 98.6& 96.8& 6.5& 96.3& 76.7& 93.7& 96.4& 97.3&
        80.1& 90.0& 86.4& 96.5& 96.5& 95.3& 98.7& 82.9&
        91.7& 81.9& 98.1& 91.3&92.1
        \tabularnewline
        
 \hline
$MCL\ (\beta=2.0)$ & 99.7& 98.0& 98.1& 97.9& 79.5& 94.7& 96.9& 97.9& 81.3&  94.2& 85.9& 98.0& 97.5& 96.1& 98.8& 84.1& 94.5& 83.5&
        98.6& 92.8 & 93.40
        \tabularnewline
    
\hline
\end{tabular}}
}
\newline
\caption{\small{Comparisons between Cross Entropy (CE), Binary Cross Entropy (BCE) and the proposed method (MCL) on VOC2007 dataset. \label{table:MCE}}}
\end{table*}
Table \ref{table:MCE} shows the effectiveness of our proposed method, MCL, over the multi-class softmax. More specifically, BCE and CE achieved mAP of $89.9\%$ and $92.1\%$, respectively, whereas MCL achieved $93.40\%$. This gain over the baseline shows the impact of contrastive learning for multi-label recognition.

\paragraph{\textbf{Parameter $\beta$:}}

\begin{figure*}[h]
\centering
        \begin{subfigure}{0.45\textwidth}
    \includegraphics[width=0.95\linewidth, height=0.15\paperheight]{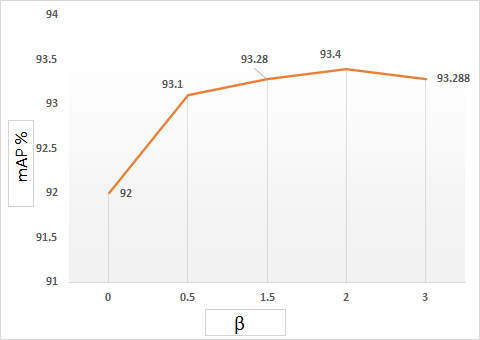}
    \caption{}
     \end{subfigure}
    \begin{subfigure}{0.45\textwidth}
            \includegraphics[width=0.95\linewidth, height=0.15\paperheight]{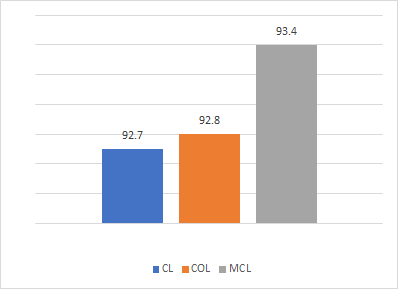}
            \caption{}
     \end{subfigure}
    \caption{\textbf{Left:} The mean Average Precision (mAP) of the proposed on VOC 2007 with different values selected for $\beta$ from $0.0$ to $3.0$. \textbf{Right:} The mean Average Precision (mAP) of the proposed method based on keeping intra-class minimization term only (\textit{i.e.} CL), inter-class maximization only (\textit{i.e.} COL) and keeping both terms (\textit{i.e.} MCL) of the multi-label contrastive loss respectively.}
    \label{fig:ablations}
\end{figure*}

The parameter $\beta$ is essential to our proposed method as it controls the level of discrepancy between classes. Therefore, we study the sensitivity of choosing a proper value for this hyper-parameter by conducting multiple experiments, where $\beta$ is chosen between $0$ and $3$. The mAP accuracies on the VOC 2007 dataset is shown in Figure \ref{fig:ablations}(a), where the AP for each class is shown in Table \ref{table:beta} for $\beta$ equals $0.5$, $1.5$ and $2.0$. It is clear that a proper choice for the value of $\beta$ affects on the model performance. Apart from that, the proposed method remains stable with various values of $\beta$.
\begin{table*}[h]
\centering
\scalebox{1}{
\resizebox{\textwidth}{!}{\begin{tabular}{|l|c|c|c|c|c|c|c|c|c|c|c|c|c|c|c|c|c|c|c|c|c|}
\hline 
Method & aero & bike & bird & boat & bottle & bus & car & cat & chair & cow & table & dog & horse & motor & person & plant & sheep & sofa & train& tv & mAP\tabularnewline
\hline 
$\beta=0.5$ & 99.5& 98.1& 98.1& 97.8& 78.4& 94.38& 96.9& 98.0&
        81.0& 94.3& 84.6& 98.0& 97.8& 96.0& 98.8& 84.1&
        94.4& 83.0& 98.6& 92.2& 93.20 
 \tabularnewline
 \hline
 $\beta=1.5$ & 99.6& 98.2& 98.3& 98.2& 80.0& 94.3& 97.1& 97.7&
        79.5& 94.5& 84.8& 97.7& 97.9& 96.4& 98.8& 83.5&
        95.7& 84.4& 98.1& 90.8 & 93.28
         \tabularnewline
 \hline
$\beta=2.0$ & 99.7& 98.0& 98.1& 97.9& 79.5& 94.7& 96.9& 97.9& 81.3&
        94.2& 85.9& 98.0& 97.5& 96.1& 98.8& 84.1& 94.5& 83.5&
        98.6& 92.8 & 93.40
        \tabularnewline
\hline
\end{tabular}}
}
\newline
\caption{\small{Comparisons between various values of $\beta$ on VOC2007 dataset. \label{table:beta}}}
\end{table*}

\begin{table*}[h]
\centering
\scalebox{1}{
\resizebox{\textwidth}{!}{\begin{tabular}{|l|c|c|c|c|c|c|c|c|c|c|c|c|c|c|c|c|c|c|c|c|c|}
\hline 
Method & aero & bike & bird & boat & bottle & bus & car & cat & chair & cow & table & dog & horse & motor & person & plant & sheep & sofa & train& tv & mAP\tabularnewline
\hline 
$\mathcal{L}_{CL}$ & 99.4& 98.1& 97.9& 98.2& 78.1& 95.5& 96.9& 97.8&
        78.98& 92.8& 81.6& 98.3& 97.1& 96.4& 98.8& 83.2&
        95.3& 81.2& 98.5& 91.8 & 92.8
 \tabularnewline
 \hline
 $\mathcal{L}_{COL}$ & 99.5& 98.1& 98.0& 97.6& 79.5& 94.2& 96.8& 97.4&
        79.5& 92.4& 84.0& 97.6& 96.6& 95.9& 98.7& 84.3&
        93.3& 82.9& 97.9& 90.2 & 92.7
 \tabularnewline
 \hline
 MCL& 99.6& 98.3& 98.0& 98.2& 78.2& 94.2& 97.0& 97.8&
        80.8& 94.9& 84.9& 97.7& 97.5& 96.6& 98.7& 85.0&
        96.2& 83.2& 98.5& 92.6 & 93.4
 \tabularnewline
\hline
\end{tabular}}
}
\newline
\caption{\small{Experiments on VOC 2007 to show the effectiveness of each component of the proposed MCL.}}
\label{table:ablation}
\end{table*}
%
\paragraph{\textbf{Intra-class minimization:}}
In this section, we investigate the effectiveness of the choosing components of the proposed Multi-label Contrastive Loss (MCL), particularly the first part of Eq. \ref{Eq:lmc}. This part is responsible for minimizing the distance between the same-class representations. By keeping only this term, Eq. \ref{Eq:lmc} will be as follows:
%
\begin{align}
\mathcal{L}_{CL} = \lce + \frac{1}{2n}\sum_{i=1}^{r}\sum_{j=1}^n{\parallel {z}^i_j - {c}_j \parallel_{2}^2}
\label{Eq:total_loss}
\end{align}

In this experiment, the second part of Eq. \ref{Eq:lmc} is ignored in order to test the impact of the intra-class variation minimization only on learning discriminative representation for different classes. Table \ref{table:ablation} ($2^{nd}$ row) shows the effectiveness of $\mathcal{L}_{CL}$, while it achieved a mAP of $92.8\%$. Although the accuracy of using $\mathcal{L}_{CL}$ is less than using total loss, it still achieving better results than the state-of-the art techniques.
\paragraph{\textbf{Inter-classes maximization:}}
We study the impact of choosing the loss function based only on the second part of Eq. \ref{Eq:lmc}, which represents the inter-class maximization. Here, we ignore the first part of the total loss function, which results in the loss function to be as follows:
\begin{align}
   \mathcal{L}_{COL} =\lce - \frac{1}{2n} \beta\sum_{i=1}^{r} \sum_{j=1}^{r}{\parallel {z}^i - {c}_j \parallel_{2}^2}, 
  \end{align}
The results in Table \ref{table:ablation} ($3^{rd}$ row) show the inter-classes dissimilarity maximization as competitive to the state-of-the art techniques and it is less than the accuracy of using the total loss function (\textit{i.e.} MCL). Also, in Figure \ref{fig:ablations} (b), a comparison between the results of using the three types of loss functions is shown.
\section{Conclusion}
Generally, learning discriminative representations is challenging for many tasks in Computer Vision. However, it becomes more challenging for multi-label recognition task due to the combinatorial nature of the input space, feature space and output space. In order to learn discriminative features for multi-label recognition, the feature space should be disentangled from many-to-many mapping into one-to-one. Then, these features can be discriminated effectively. In this paper, we have proposed a novel method to achieve the transition from many-to-many mapping to one-to-one in the feature space with the help of the annotated bounding boxes. Then, a contrastive loss, namely multi-label contrastive loss (MCL), is proposed to discriminate the features and alleviate the unbalanced data distributions. Experimental results were conducted to demonstrate the prominence of the proposed method. As shown quantitatively and qualitatively, our proposed method was able to outperform all the baselines without any additional supervision, external embedding or involving complex network architectures. However, MCL inherits some limitations of hyper-sphere loss functions such as time concerns because it compares all the batch items individually and thus it behaves better in small datasets. Also, it behaves better in the case of small number of classes (\textit{e.g.,} VOC PASCAL) rather than large number ones (COCO). In the future, we plan to study these two limitations and propose generic solutions for them. 

\small{
\bibliographystyle{cas-model2-names}
\bibliography{main}
}

\end{document}